\documentclass[10pt,letterpaper]{article}

\usepackage{cogsci}
\usepackage{hyperref} 

\cogscifinalcopy 

\usepackage{pslatex}
\usepackage{apacite}
\usepackage{float} 

\usepackage{url}

\usepackage{subcaption}
\usepackage{graphicx}
\usepackage{booktabs}
\usepackage{multirow}
\usepackage{array}
\usepackage{caption}
\newcolumntype{C}[1]{>{\centering\arraybackslash}p{#1}}

\title{CV-Probes: Studying the interplay of visual grounding and social knowledge in verb understanding}
 
\author{{\large \bf Ivana Benova (ivana.benova@kinit.sk)} \\
  Faculty of Informatics, Brno University of Technology, Czech Republic  \\
  Kempelen Institute of Intelligent Technologies, Slovakia
  \AND {\large \bf Michal Gregor (michal.gregor@kinit.sk)} \\
 Kempelen Institute of Intelligent Technologies, Slovakia
  \AND {\large \bf Albert Gatt (a.gatt@uu.nl) } \\
  Utrecht University, Utrecht, Netherlands}

\begin{document}

\maketitle

\begin{abstract}
How do vision-language (VL) transformer models ground verb phrases and do they integrate contextual and world knowledge in this process? We introduce the CV-Probes dataset, containing image-caption pairs involving verb phrases that require both social knowledge and visual context to interpret (e.g., `beg'), as well as pairs involving verb phrases that can be grounded based on information directly available in the image (e.g., ``sit"). We show that VL models struggle to ground VPs that are strongly context-dependent. Further analysis using explainable AI techniques shows that such models may not pay sufficient attention to the verb token in the captions. Our results suggest a need for improved methodologies in VL model training and evaluation. The code and dataset will be available \url{https://github.com/ivana-13/CV-Probes}.

\end{abstract}

\section{Introduction}
\begin{figure}[t]
    \centering
    \begin{subfigure}[t]{0.45\linewidth}
        \centering
        \includegraphics[width=3.67cm]{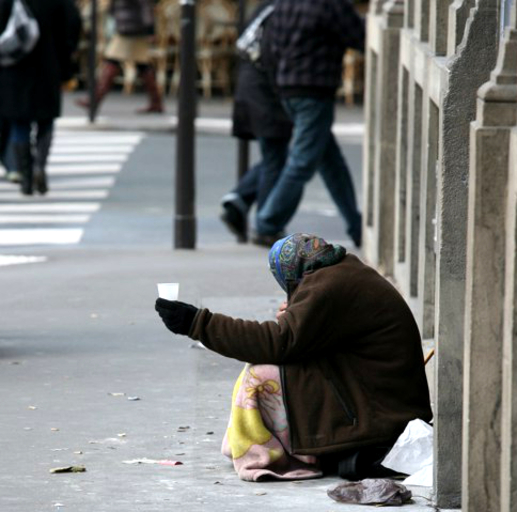}
        \caption{\small A woman begs for money in the street. (A)\\A woman sits in the street with a cup in her hand. (B)}
        \label{fig:image_a}
        \vfill
    \end{subfigure}
    \hfill
    \begin{subfigure}[t]{0.45\linewidth}
        \centering
        \includegraphics[width=3.23cm]{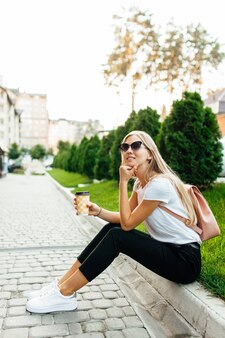}
        \caption{\small A woman sits in the street with a cup in her hand. (B)}
        \label{fig:image_b}
    \end{subfigure}
    \vspace*{-0.5em}
    \caption{Example from the CV-Probes dataset. While image \ref{fig:image_a} can be described by both a socially grounded (A) and physically grounded caption (B), only the latter (B) applies to image \ref{fig:image_b}.}    
    \label{fig:both_images}
    \vspace*{-1em}
\end{figure}

The meaning of natural language expressions has two important dimensions. On the one hand, meaning can be computed based on lexical and compositional semantics. On the other, meaning also relies on conceptual and world knowledge. This distinction broadly corresponds to that between {\em formal} linguistic competence---the ability to produce and understand expressions correctly based on their form---and {\em functional} competence, the ability to produce utterances that satisfy communicative goals and interpret them based on what we know about the world \cite{mahowald_dissociating_2024}. When computing meaning, a natural or artificial system also connects linguistic symbols to the non-linguistic world, a challenge sometimes referred to as symbol grounding \cite{Harnad1990}. In AI, the extent to which Large Language Models (LLMs) are (or should be) able to connect natural language symbols to non-linguistic data remains a highly debated question \cite{Bender2020,Bisk2020,pavlick_symbols_2023,mandelkern_language_2024,leivada_sentence_2024}. Functional competence plays a role here, too. Consider determining whether a caption correctly describes one of the visual scenes in Figure~\ref{fig:both_images}. Both scenes involve a seated woman holding a cup in her hand. A description involving these details can be verified against an image by linking specific predicates to image regions. On the other hand, most English speakers would agree that only one of the images of seated women also involves the action of {\em begging}. Determining this involves reasoning about the woman's intentions, the social significance of her actions, and other visual cues in the image. 
As \citeA{mahowald_dissociating_2024} argue, reasoning with social and world knowledge is also part of our functional linguistic competence. 

In the field of multimodal learning, transformer-based models have significantly advanced the integration of visual and textual data in an attempt to partially address the grounding problem \cite{tan2019lxmert,li2021align,alayrac-flamingo-2022,zeng-multi-grained-2022,li2023blip,liu-visual-2023,dai-instructblip-2023,deitke_molmo_2024}. 
Various benchmarks have been developed to assess such models' ability to ground fine-grained linguistic phenomena, including verbs \cite{hendricks2021probing, bevnova2024beyond}, counting \cite{parcalabescu2020seeing}, spatial relations \cite{kamath-whats-2023}, and word order \cite{thrush2022winoground,chen-bla-2023,ray-cola-2023,ma-crepe-2023}.

For VLMs, verb phrases (VPs) pose a significant challenge. Studies have shown that they
struggle with VP-centered tasks such as distinguishing cases involving swapping of verbs ({\em the woman \underline{fired/held} the gun}) or arguments ({\em the woman shouted at the man} vs. {\em the man shouted at the woman}) \cite{parcalabescu2021valse,parcalabescu_measuring_2024}. Similarly, video-language models have difficulty grounding expressions involving temporal or situational elements, including VPs \cite{kesen2024vilma,xiao_can_2024}. Some of these results suggest a lack of formal competence (e.g., compositional interpretation of a verb and its arguments); others veer towards the functional (the difference between holding and firing in part relies on subtle visual cues and world knowledge).

This paper focuses on grounding VPs.
We analyze VLM performance on image-text pairs like those in Figure \ref{fig:both_images}. In each pair, the images have several visual features in common (e.g. the position of the subject in Figure~\ref{fig:both_images}, the extended arm holding a cup, the outdoor setting). These explicit, physical features support the use of a verb (like {\em sit}) for both images. On the other hand, certain other features (e.g. dress, type of cup, posture) warrant different inferences about the subject's intention and/or purpose and rely on social and world knowledge. For example, only the person in Figure \ref{fig:image_a} is likely to be \textit{begging}. Our question is to what extent pretrained VLMs distinguish verbs like {\em sit} and {\em beg} by grounding them in the relevant features of the visual context, leveraging additional social knowledge to solve the grounding problem where relevant.
%
To this end, we develop the \textbf{C}ontextual \textbf{V}erb-Probes (CV-Probes) dataset.
We follow previous work that relies on the image-text matching capabilities of VLMs. However, this method has known limitations in revealing a model's grounding abilities \cite{bevnova2024beyond}. Hence, we further rely on MM-SHAP \cite{parcalabescu2022mm}, a performance-agnostic metric to quantify and interpret the contribution of individual tokens and modalities in VL models—to quantify the contribution of each input token (both visual and textual) to the model's predictions, with particular emphasis on the verbs in the captions.


\section{Related Work}

\paragraph{Vision-Language Models (VLMs)} are typically trained on large datasets of image-text pairs \cite{changpinyo2021conceptual,schuhmann2021laion}, and combine a visual backbone (e.g., vision transformers \cite{Dosovitskiy2020}) with a textual encoder or decoder. Early VLMs, based on BERT \cite{Devlin2019}, used single- or dual-stream architectures \cite{tan2019lxmert, Li2019, Lu2019,bugliarello2021multimodal}. Dual-encoders like CLIP \cite{radford2021learning} and ALIGN project visual and linguistic features into a common space. Recent models, such as Flamingo \cite{alayrac-flamingo-2022}, BLIP2 \cite{li2023blip} and LlaVA-NeXT \cite{liu2024llavanext}, use frozen visual and textual modules like LLMs linked by intermediate networks. Training objectives include masked or autoregressive language modeling, contrastive learning; and image-text matching. Instruction-tuned VLMs \cite{liu-visual-2023,dai-instructblip-2023, openai_gpt4o} extend these capabilities with interactive features. 

\paragraph{Fine-grained benchmarks} Several benchmarks study the abilities of VLMs to ground fine-grained linguistic phenomena and learn compositional language representations  \cite{thrush2022winoground, ma-crepe-2023,kamath-whats-2023, kamath_hard_2024}. 
An important paradigm is foiling \cite{shekhar2017foil}, whereby a dataset with images and corresponding texts (`positive' cases) is manipulated by changing essential parts of the caption, resulting in a {\em foil}, which is no longer true of the image. While CV-Probes is not strictly a foil-based benchmark, our experiments rely on the ability of a model to distinguish captions that differ in a specific linguistic phenomenon. In the same spirit, and closest to our work, are benchmarks with an explicit focus on verb phrase (VP) grounding, such as SVO-Probes \cite{hendricks2021probing} and parts of VALSE \cite{parcalabescu2021valse} and ViLMA \cite{kesen2024vilma}, which use a foiling-based method to probe VLM verb understanding, among other things. Focusing on fine-grained linguistic phenomena ultimately contributes to our broader understanding of VLMs' limits on grounded compositional reasoning tasks. 
Unlike previous work, CV-Probes explicitly focuses on VP interpretation in cases where there are both shared and distinct elements of the visual context.

\section{Dataset}
The CV-Probes dataset is designed with the following rationale. We start from images depicting actions that require reasoning about a person's intentions, based on a combination of visual context and social or world knowledge (caption A in Figure~\ref{fig:image_a}). We refer to the verbs describing these images as `socially grounded' (SG), since understanding these verbs requires reasoning about social factors and intentions, for which some visual features serve as cues. We contrast these verbs with alternatives that can describe the same images (caption B in Figure~\ref{fig:image_a}) based on visual features, but which do not require reasoning about intentions beyond what is literally depicted. We refer to these as `physically grounded' (PG).
We further pair these image-caption pairs with additional images to which only the PG caption can apply (Figure~\ref{fig:image_b}). Crucially, these images are visually similar to the ones depicting SG actions (both images in Figure~\ref{fig:both_images} depict females seated outdoors). A VLM should assign a high image-text matching probability to both captions (A and B) for the image in Figure~\ref{fig:image_a}, but only to the PG caption (B) for the image in Figure~\ref{fig:image_b}. Abusing terminology somewhat, we will for the sake of brevity refer to images describable by captions containing socially grounded VPs as `SG images' and to images not describable by such captions as `PG images.'

\subsection{Data selection and preprocessing}
 We leveraged the ImSitu dataset \cite{yatskar2016}, a comprehensive resource for situation recognition tasks. ImSitu provided a foundation of 504 unique verbs and corresponding images, each encapsulating recognizable activities in their corresponding images. We selected 27 verbs that strongly rely on social and world knowledge to interpret them in a visual scene. The list of all SG verbs is in \tablename~\ref{List_of_verbs}.

   \begin{table}[ht!]
   \footnotesize
\caption{Manually selected verbs for socially grounded (SG) captions from ImSitu dataset.}
\vspace*{-0.5em}
\begin{tabular}{p{8.2cm}} \hline
admire, apprehend, autograph, baptize, beg, brows, buy, celebrate, chase, cheerlead, coach, compete, congregate, count, educate, exercise, frisk, guard, hitchhike, hunt, interrogate, interview, pray, protest, race, study, vote \\ \hline
\end{tabular}
\label{List_of_verbs}
\end{table}

Each verb was paired with five different images from ImSitu. Verbs were further mapped to their FrameNet \cite{ruppenhofer2016framenet} frames. ImSitu specifies the roles or arguments of each verb; we crafted three SG descriptions per verb by mapping these arguments to the corresponding FrameNet template. 
These captions aimed to provide concise yet informative descriptions.
We used ChatGPT 3.5 to make the captions more fluent.
For consistency, all verbs are in the present tense and all articles are indefinite.
We used the GRUEN pre-trained model \cite{Zhu2020} to further score the captions for grammaticality and selected the caption with the highest score as the SG caption for each image. The average score of all captions was $0.793 \pm 0.017$ (the top GRUEN score is $1$), while the average score of the best-selected captions was $0.848 \pm 0.021$. 

 \subsection{Physically grounded captions and images}
 Four independent annotators further annotated each image. The annotators were volunteers, non-native but fluent English speakers with a tertiary education. They were tasked with generating PG captions that reflected the depicted scenario. Annotators were instructed to describe what they saw, avoiding captions relying on world knowledge and inference. To facilitate this, they were explicitly
 provided with the SG captions and asked to avoid the SG verb in their new captions. Annotations were collected using the Doccano \cite{doccano} platform; we had 20 volunteers.
 Subsequently, annotations underwent a review process by two of the authors to ensure descriptive fidelity. Annotations that used SG verbs in violation of the annotator instructions were removed. This iterative refinement process resulted in a curated dataset comprising one SG caption and one or more descriptive PG captions for each image.

We paired the PG captions with new images using the CLIP model \cite{radford2021learning} to perform image retrieval from the LAION dataset \cite{schuhmann2021laion}. Given a PG caption, we retrieved image candidates and manually checked that the images corresponded only to the PG captions, but not the SG ones (Figure~\ref{fig:image_b}). 

\begin{table}[t!]
\caption{CV-Probes dataset statistics}
\vspace*{-0.5em}
    \centering
    \begin{tabular}{l|c} \toprule
    image-caption pairs & \# \\ \midrule
    socially grounded pairs & 117 \\
    physically grounded pairs & 171 \\ \bottomrule
    \end{tabular}
    \label{tab:datastats}
    \vspace*{-1.2em}
\end{table}

\subsection{Caption simplification}
CV-Probes focuses on model's ability to match captions to images based on verb phrases.
However, captions may contain additional information, often in syntactic arguments and adjuncts, which models may rely on during image-text matching. Thus, we manually created a simplified version of the captions, consisting exclusively of simple declarative sentences with a subject, verb, and object, but no additional modifiers (such as `in the street' in Figure \ref{fig:both_images}). Table \ref{tab:datastats} gives the statistics for the final CV-Probes dataset.

\subsection{Human validation}
\begin{figure}[h!]
    \centering
    \begin{subfigure}[h!]{0.45\linewidth}
        \centering
        \includegraphics[width=3.1cm]{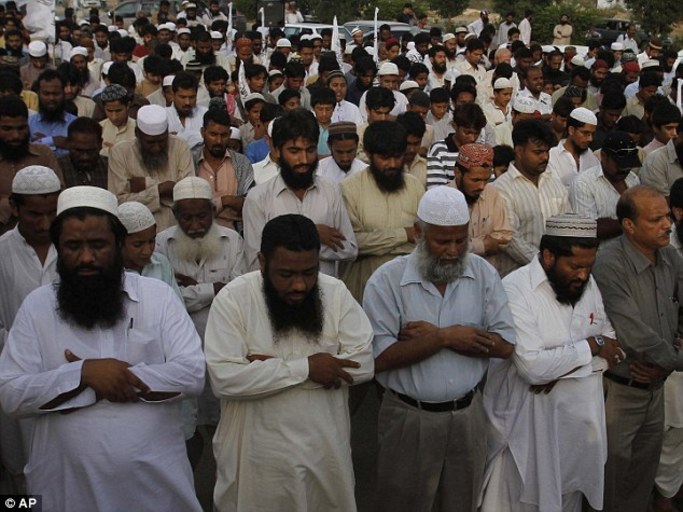}
        \caption{\small Some people pray. (A)\\A group of men stand with crossed arms. (B)}
        \label{fig:image_aa}
        \vfill
    \end{subfigure}
    \hfill
    \begin{subfigure}[h!]{0.45\linewidth}
        \centering
        \includegraphics[width=3.1cm]{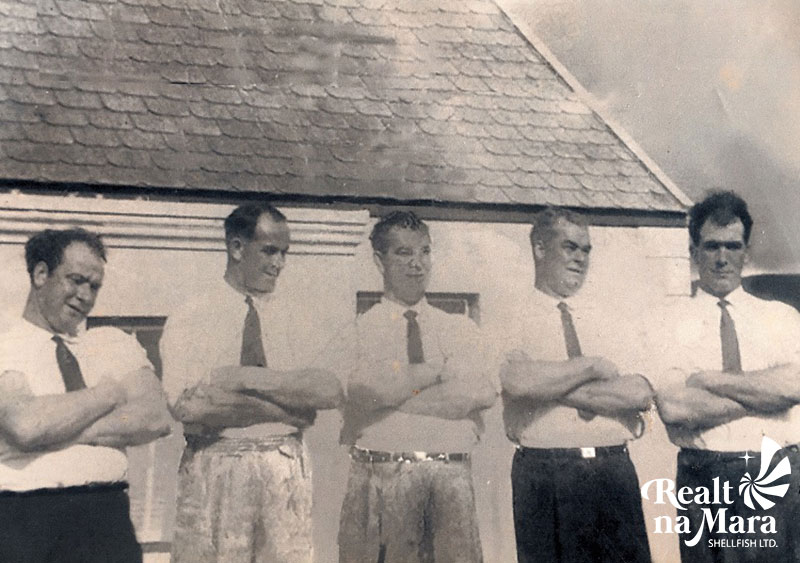}
        \caption{\small A group of men stand with crossed arms. (B)}
        \label{fig:image_bb}
    \end{subfigure}
    \vspace*{-0.5em}
    \caption{Additional example from the CV-Probes dataset. While image \ref{fig:image_aa} can be described by both a socially grounded (A) and a physically grounded caption (B), only the latter (B) applies to image \ref{fig:image_bb}.}    
    \label{fig:both_images_2}
    \vspace*{-1em}
\end{figure}

We performed a second round of human validation on the simplified captions only, crowd-sourcing annotations via the Prolific platform,\footnote{https://www.prolific.com/} with native speakers of English. Annotations were done over two separate sessions. In the first one, each annotator was presented with two images (SG and PG) and one caption (either SG or PG), and they were asked to choose whether the caption applies to both images, only one of them or neither. Three annotators evaluated each triplet. 6 annotators participated in this study. For triplets where at least 2 annotators chose different answers than the presumed true label, we manually improved the caption. The annotators were compensated by $9\pounds$ per hour of work. The resulting filtered dataset was validated in a second session to obtain human judgments against which we compare AI models.
Each image-caption pair was evaluated by 3 annotators (24 in total). A further example of from the dataset can be seen in \ref{fig:both_images_2}.

\section{Model Performance on CV-Probes }
\label{IMT}
To evaluate performance on the CV-Probes dataset, we leverage the capability of pretrained VLMs to perform image-text matching. For VLMs with an image-text matching head, we treat the task as a binary classification: given an image-text pair $\langle I, T \rangle$, the task is to determine the probability that $T$ describes $I$. For a matching pair, a model response is considered correct if it assigns $p(1 \vert I, T) > p(0 \vert I, T) $, where $1$ indicates a positive match. Similarly, for a non-matching pair, a model is correct if $p(0 \vert I, T) > p(1 \vert I, T) $. We evaluate each image-caption pair separately. We do not explicitly
compare matching and non-matching pairs, since (a) for SG images, both SG and PG captions apply;
and (b) we consider our evaluation strategy as ultimately very lenient, since we only consider whether the probability of a correct  match exceeds that of an incorrect match.

Our evaluation involved several models that are representative of different neural architectures. We introduce them briefly here.
LXMERT~\cite{tan2019lxmert} is a dual-stream architecture that integrates visual features extracted using a Faster R-CNN visual backbone with linguistic features extracted using a BERT architecture. 
ALBEF \cite{li2021align} encodes visual features with a vision transformer and linguistic features with the first six layers of BERT, using the last six BERT layers for multimodal fusion via cross-attention. 
BLIP~\cite{li2022blip} employs a visual transformer and a BERT language model, integrating these through a Multimodal Mixture of Encoder-Decoder (MED) architecture. 
In contrast, BLIP2~\cite{li2023blip} introduces a Querying Transformer (Q-Former) to bridge the gap between a frozen image encoder and a frozen large language model (LLM). 
FLAVA~\cite{singh2022flava} comprises a visual and textual transformer, each trained on modality-specific data before being integrated through a shared multimodal encoder. 
LLaVA-NeXT~\cite{liu2024llavanext} uses a pretrained visual backbone with a large language model decoder, with pretraining stages including image-text matching and image-to-text generation. 
We also compare the results of these models with GPT-4o \cite{openai_gpt4o}, a state-of-the-art model whose weights and architecture are not available. Both LLaVa-NeXT and GPT-4o
were evaluated using a prompt asking whether a given caption matches an image. Hence, these models generate a response (match or non-match), while the other models explicitly compute the probability of an image-text match. Since GPT-4o is a closed model, we only report performance results for comparison purposes, reserving the in-depth analysis for other models.

\subsection{Results}
We report model accuracy on both the original (Table~\ref{ITM_original}) and the simplified captions (Table~\ref{ITM_simlified}) in CV-Probes. 

\begin{table*}[t!]
\caption{Image-text matching accuracy (\%) on the original captions. SG: socially grounded; PG: physically grounded.}
\vspace*{-0.5em}
\small
\centering
\begin{tabular}{lC{4.5em}C{5em}C{4.5em}C{5em}C{6.5em}}
\toprule
    \multicolumn{1}{c}{{\em Images}:}  & \multicolumn{2}{c}{\bf SG (cf. Fig~\ref{fig:image_a})} & \multicolumn{2}{c}{\bf PG (cf. Fig.~\ref{fig:image_b})}
\\
    \multicolumn{1}{c}{{\em Captions}:} &
        {\em SG} &
        {\em PG} &
        {\em SG} &
        {\em PG} &
        \multicolumn{1}{C{6.5em}}{Harmonic mean}
\\    
    \cmidrule{2-3} \cmidrule{4-6}
    LXMERT & \multicolumn{1}{c}{26.50} & 37.79 & \multicolumn{1}{c}{83.63} & 41.52 & 39.90
\\ 
    ALBEF & \multicolumn{1}{c}{63.25} & 70.18 & \multicolumn{1}{c}{74.85} & 92.98 & 73.84
\\  
    BLIP & \multicolumn{1}{c}{78.63} & 75.58 & \multicolumn{1}{c}{78.36} & 92.40 & \textbf{80.75}
\\  
    BLIP 2 & \multicolumn{1}{c}{62.39} & 55.23 & \multicolumn{1}{c}{\textbf{89.47}} & 80.70 & 69.32
\\  
    FLAVA & \multicolumn{1}{c}{\textbf{98.29}} & \textbf{87.13} & \multicolumn{1}{c}{22.22} & \textbf{95.91} & 51.89
\\ \cmidrule{2-3} \cmidrule{4-6} 
    LlaVA-NeXT & 94.87 & 95.91 & 61.63 & 98.84 & 84.55 \\
    GPT-4o & 88.89 & 86.55 & 85.38 & 87.13 & 86.97 \\
\bottomrule
\end{tabular}
\vspace*{-1em}
\label{ITM_original}

\end{table*}

\begin{table*}[t!]
\caption{Image-text matching accuracy (\%) on the simplified captions. SG: socially grounded; PG: physically grounded.}
\vspace*{-0.5em}
\small
\centering
\begin{tabular}{lC{4.5em}C{5em}C{4.5em}C{5em}C{4em}}
\toprule
    \multicolumn{1}{c}{{\em Images}:} & \multicolumn{2}{c}{\bf SG (cf. Fig~\ref{fig:image_a})} & \multicolumn{2}{c}{\bf PG (cf. Fig.~\ref{fig:image_b})}
\\ 
    \multicolumn{1}{c}{{\em Captions}:} &
        {\em SG} &
        {\em PG} &
        {\em SG} &
        {\em PG} &
        \multicolumn{1}{C{6.5em}}{Harmonic mean}
\\     \cmidrule{2-3} \cmidrule{4-6} 
    LXMERT & 29.91 & 41.86 & 77.78 & 38.60 & 41.63
\\ 
    ALBEF & 51.28 & 53.80 & 71.93 & 78.36 & 61.77
\\ 
    BLIP & 75.21 & 59.88 & 74.27 & 84.80 & \textbf{72.39}
\\ 
    BLIP 2 & 47.01 & 36.63 & \textbf{91.23} & 58.48 & 52.20
\\ 
    FLAVA & \textbf{96.58} & \textbf{88.89} & 12.87 & \textbf{94.15} & 36.39
    \\ \cmidrule{2-3} \cmidrule{4-6} 
    LlaVA-NeXT & 94.02 & 97.66 & 43.60 & 100.00 & 74.33 \\
    GPT-4o & 91.45 & 91.81 & 78.36 & 94.15 & 88.48 \\ \cmidrule{2-3} \cmidrule{4-6} 
    Human & 98.29 & 97.66 & 79.53 & 92.40 & 91.31
\\ \bottomrule
\end{tabular}
\label{ITM_simlified}
\vspace*{-1em}
\end{table*}

In \tablename~\ref{ITM_original}, the highest accuracy for matching pairs was obtained by  FLAVA. For SG images, FLAVA is at or close to the maximum performance ($98.29\%$ and $87.13\%$ for SG and PG captions respectively). For matching PG images and captions, it reaches an accuracy of $95.91\%$, surpassing all other models except the decoder-based LlaVA-NeXT in this regard. Conversely, LXMERT, ALBEF, and BLIP2 exhibited comparatively lower accuracy, and usually below chance (LXMERT at $26.50\%$) to $78.6\%$ (BLIP) for images depicting context-dependent actions. 

However, the most important results are in column 3, providing the accuracy for pairs of SG captions paired with PG images. By construction, these are non-matching cases. Here, FLAVA's performance drops significantly to below chance levels. This suggests a positive bias in the model and a tendency to assign high probabilities to image-text pairs, even in non-matching cases. Similar observations are made for specific models in SVO-Probes by \cite{hendricks2021probing}. 
On the other hand, the best performance in this category is achieved by BLIP2 with an accuracy of $89.47\%$. 
We note that BLIP2 performance also exceeds that of LlaVA-NeXT and GPT-4o on these non-matching cases.

The same trends are broadly seen in \tablename~\ref{ITM_simlified}, which contains results for the simplified captions. We observe that simplification sometimes causes a model to change its prediction. One possibility is that the process of simplification, which, for example, removes adjunct phrases in the captions (such as `on the street' in Figure~\ref{fig:both_images}), leaves models with less signal in the textual modality to boost the probability of a match. This, in turn, may suggest that models rely on the non-verb parts of the captions more, an issue we return to in the following section.

A specific case is the evaluation of the LLaVA-NeXT and GPT-4o. LLaVA-NeXT performs very well on matching pairs, but its performance on non-matching pairs is close to random guessing and shows a significant drop for simplified captions (compare column 3, Tables~\ref{ITM_original} and \ref{ITM_simlified}). The performance of GPT-4o is however very close to human performance, lacking mostly for matching pairs. However, as these models are generative, we can not compare the results with the rest of the models that were evaluated using image-text matching and we also can not use MM-SHAP evaluation as we do in the next section with the rest of the models.

It is worth considering possible reasons why BLIP2 performs much better than other models, including FLAVA, on the key condition. One might be architectural: while all other models rely on a single token in the image-text matching head (usually, a {\tt CLS} or {\tt Encode} token),  BLIP2 has 32 learnable query embeddings as input to the query transformer. The queries interact with each other through self-attention layers and interact with frozen image features through cross-attention layers. Queries are projected via a linear classifier, and the resulting logits across all queries are averaged to get the matching score. Another distinguishing feature of BLIP2 is that it is pretrained with image-grounded text generation. Finally, BLIP2 differs from other models with a vision transformer visual backbone in that it uses ViT-g as an image encoder, compared to ViT-B or ViT-L used in other models.
As we noted, FLAVA exhibits an overall positive bias and fails in those cases where a SG caption does not match an image. FLAVA has unimodal backbones for vision and language, and incorporating unimodal pretraining (as well as other losses) improves the performance on vision tasks, NLP tasks, and multimodal tasks \cite{singh2022flava}. On the other hand, image-text matching in FLAVA did not involve mining for (hard) negatives. This may underlie the positive bias: the model may predict a match based on, for example, only matching entities (`woman') or locations (`street').

\begin{table*}[t!]
\footnotesize
\caption{Pearson correlation between human annotations and model classification in CV-Probes. Non-significant correlations (at $p \geq 0.05)$ are in italics.}
\vspace*{-0.5em}
\small
\centering
\begin{tabular}{lC{4.5em}C{5em}C{4.5em}C{5em}}
\toprule
    \multicolumn{1}{c}{{\em Images}:}  & \multicolumn{2}{c}{\bf SG (cf. Fig~\ref{fig:image_a})} & \multicolumn{2}{c}{\bf PG (cf. Fig.~\ref{fig:image_b})}
\\
    \multicolumn{1}{c}{{\em Captions}:} &
        {\em SG} &
        {\em PG} &
        {\em SG} &
        {\em PG} 
\\    
    \cmidrule{2-3} \cmidrule{4-5}
    LXMERT & \multicolumn{1}{c}{0.248} & {\it 0.04} & \multicolumn{1}{c}{0.209} &  {\it 0.104}
\\ 
    ALBEF & \multicolumn{1}{c}{0.187} & 0.2 & \multicolumn{1}{c}{{\it 0.143}} & {\it 0.093}
\\  
    BLIP & \multicolumn{1}{c}{{\it 0.116}} & 0.267 & \multicolumn{1}{c}{0.181} & {\it 0.144}
\\  
    BLIP 2 & \multicolumn{1}{c}{{\it 0.149}} &  0.201 & \multicolumn{1}{c}{{\it 0.142}} & 0.167
\\  
    FLAVA & \multicolumn{1}{c}{{\it -0.006}} & {\it 0.044} & \multicolumn{1}{c}{{\it 0.049}} & {\it 0.036}
\\  \cmidrule{2-3} \cmidrule{4-5} LlaVA-NeXT & {\it 0.098} & {\it -0.086} & {\it 0.023} & {\it -0.073} \\

\bottomrule
\end{tabular}
\label{ITM_correlation}
\vspace*{-1em}
\end{table*}

In Table \ref{ITM_correlation}, we present a Pearson correlation between the probabilities of image-caption matches calculated by the models and the votes obtained by human annotations in our second round (see {\em Dataset} above). While almost all correlations are positive, most have a low coefficient, and many are not significantly different from 0. Thus, VLM predictions do not correlate strongly with how humans evaluate image-caption matches in CV-Probes. Unfortunately we can not calculate correlation between human prediction score and model probability for GPT-4o as logits are not readily accessible via the GPT-4o multimodal API.

\section{Do Models Rely on Verbs?}
\label{MMShap}
In this section, we focus on token-level explanations for the image-text matching results. In particular, we ask to what extent models rely on verb tokens to compute an image-text matching probability. For this purpose, we use MM-SHAP \cite{parcalabescu2022mm}, an adaptation of SHAP \cite{lundberg2017unified}, which is an approximation of Shapley values \cite{shapley1953value}. Shapley values are a game-theoretic formalism to compute the contributions of individual players in cooperative games, given a numerical outcome or score. For a given player, the idea is to compare the contribution of each possible coalition of players when that player is included versus when they are not. For AI model explanations, we consider input features to be the `players'. Since the number of possible coalitions in a Shapley game is exponential, approximations such as SHAP provide a compute-optimal solution. Crucially, SHAP explanations are model-agnostic, making it possible to compare them across models.

MM-SHAP is designed for a VLM with $n_T$ text and $n_I$ image tokens.\footnote{Image tokens are defined as superpixels or image patches.} The textual $\phi_T$ and the image contribution $\phi_I$ towards a prediction are defined as the sum of the absolute SHAP values of textual and visual tokens, respectively:

\begin{equation}
    \phi _{T} = \sum_j^{n_T} |\phi_j| \quad ; \quad \phi_{I} = \sum_j^{n_I} |\phi_j|
\end{equation}

MM-SHAP provides one score per modality, defined as the proportion of the overall SHAP total due to that modality:
\begin{equation}
      T-SHAP = \frac{\phi_T}{\phi_T+\phi_I}; \quad V-SHAP = \frac{\phi_I}{\phi_T+\phi_I}
\end{equation}

This formulation offers insight into potential unimodal collapse, whereby a model relies extensively on one modality to the detriment of the other \cite{Hessel2020, Frank2021,parcalabescu2022mm}.

\begin{table*}[t!]
\caption{Average T-SHAP scores for the caption (overall) and average for the verb only, with proportion of overall SHAP attribution for the verb for 50 samples. Scores are shown separately for the case where the model predicts that the image and caption match ($p > 0.5$) and do not match. SG: socially grounded; PG: physically grounded.}
\tiny
    \centering
    \begin{tabular}{rr|cccc|cccc|cccc}
    \toprule
    & & \multicolumn{4}{c|}{BLIP} & \multicolumn{4}{c|}{BLIP2} & \multicolumn{4}{c}{FLAVA} \\
    \midrule
        \multicolumn{2}{r|}{{\em Images}:} &
            \multicolumn{2}{c}{
                \bf SG (cf. Fig~\ref{fig:image_a})
            } & \multicolumn{2}{c|}{
                \bf PG (cf. Fig~\ref{fig:image_b})
            } &  \multicolumn{2}{c}{
                \bf SG (cf. Fig~\ref{fig:image_a})
            } & \multicolumn{2}{c|}{
                \bf PG (cf. Fig~\ref{fig:image_b})
            } & \multicolumn{2}{c}{
                \bf SG (cf. Fig~\ref{fig:image_a})
            } & \multicolumn{2}{c}{
                \bf PG (cf. Fig~\ref{fig:image_b})
            }\\
        
        \multicolumn{2}{r|}{{\em Captions}:} &
            {\em SG} &
            {\em PG} &
            {\em SG} &
            {\em PG} &
            {\em SG} &
            {\em PG} &
            {\em SG} &
            {\em PG} &
            {\em SG} &
            {\em PG} &
            {\em SG} &
            {\em PG}\\
        
    \cmidrule{3-4} \cmidrule{5-6} \cmidrule{7-10} \cmidrule{11-14}
    
        \multirow{3}{*}{\bf Match} & {\em Overall (\%)} & 89.44 & 85.41 & 88.47 & 88.46 & 77.29 & 70.85 & 65.21 & 70.75 & 45.65 & 45.39 & 45.87 & 45.78 \\
                     & {\em Verb} & 0.3790 & 0.1596 & {\bf 0.3047} &  0.1586 & 0.4266 & 0.1681 & {\bf 0.1294} & 0.0877 & 0.1236 & 0.0871 & {\bf 0.1521} & 0.0999\\
                     & {\em Verb (\%)} & 47.10 & 21.26 & 42.18 & 20.84 & 60.31 & 22.50 & 32.14 & 15.96 & 32.28 & 21.33 & 36.22 & 25.93\\
                     
    \cmidrule{3-4} \cmidrule{5-6} \cmidrule{7-10} \cmidrule{11-14}
    
        \multirow{3}{*}{\bf Non-match} & {\em Overall (\%)} & 75.20 & 77.39 & 72.64 & 81.58 & 60.99 & 56.46 & 56.10  & 52.24 & 46.66 & 47.73 & 45.42 & 52.83 \\
                     & {\em Verb} & 0.1228 & 0.0119 & {\bf -0.0323} & 0.1586 &  0.1381 & -0.0174 & {\bf -0.0131} & 0.0322 &  0.0592 & 0.0273 & {\bf 0.1394} & 0.2555\\
                     & {\em Verb (\%)} & 24.92 & 16.47 & 35.83 & 20.48 &  24.44 & 20.07 & 20.21 & 13.07 &  18.12 & 11.38 & 21.88 & 24.66\\
    \bottomrule
    \end{tabular}
    \label{BLIP}
    \vspace*{-1em}
\end{table*}

\subsection{Results}
For reasons of space, we report results on a selection of models, focusing primarily on BLIP, which obtained the highest harmonic mean overall conditions in Tables~\ref{ITM_original} and \ref{ITM_simlified}, as well as BLIP2 and FLAVA.
Table \ref{BLIP}  reports results for the three models computed over 50 samples from CV-Probes. We use simplified captions to reduce the chance that attributions arise from text tokens that do not belong to the VP or its arguments.
We report overall T-SHAP values and the contribution of the individual verb token towards match prediction. In addition, we estimate the percentage of the overall T-SHAP score that is due to the verb. We take this as an indicator of the overall importance of the verb token in accounting for the model predictions. 

The T-SHAP values for BLIP suggest that this model relies heavily on the textual modality, far more than on the visual modality, raising the possibility of unimodal collapse (T-SHAP $\gg$ 50\%). This suggests the model exploits text biases, to some extent reducing to a unimodal model for this task. Notably, the verb contribution for PG images with SG captions should lead to a non-match prediction. However, BLIP shows a high average verb attribution ($0.3047$) when incorrectly predicting a match, but the verb attribution is almost zero ($-0.0323$) when correctly predicting a non-match. This implies that correct predictions are not influenced by verb tokens, whereas incorrect predictions are.

The BLIP2 and FLAVA models are more evenly balanced between visual and textual modalities. For FLAVA in particular, T-SHAP overall is close to 50\%, perhaps unsurprisingly, given its architecture. For BLIP2, verb attributions are high for match predictions ($0.1294$) and close to zero for non-match predictions ($-0.0131$) in the key condition, indicating that verb tokens do not play a correct role in image-text matching task. For FLAVA, the second highest average verb attributions occur with PG images with SG captions. This suggests the model 
has a poor grounding ability concerning SG verbs: these contribute {\em positively} overall to both match and non-match predictions in this non-matching context.

Based on these results, we conclude that these models struggle to ground SG and PG verb phrases effectively.

\section{Conclusion}
\label{conclusion}
The ability to ground linguistic symbols in the non-linguistic world of perception and experience relies on both lexico-syntactic knowledge (part of our formal linguistic competence) and world knowledge (part of functional competence). World knowledge, in particular, mediates our ability to describe scenes and actions beyond their literal composition, by recognizing actions and intentions. In this sense, the ability to ground descriptions of actions and interpret verb phrases relies on social and world knowledge to different degrees.

In this study, we evaluated the ability of several vision-language (VL) models to ground socially grounded (`beg', `pray') and physically grounded (`sit', `stand') verb phrases within the context of image-text matching tasks. One of our primary contributions is a novel test dataset, CV-Probes, which pairs images with descriptions that rely on social and world knowledge to different degrees, with an emphasis on verb phrase grounding.

We find that VL models exhibit varying abilities to ground SG captions describing activities to visual data and tend to fail on the critical test case, where a caption with a SG verb phrase is paired with an image that does not depict such an action but is visually similar to its SG counterpart. 

We further analyzed models in a performance-agnostic way using Shapley values, focusing specifically on the contribution of verb tokens to the model's predictions. 
The results indicated that models struggled with grounding SG verb phrases. For BLIP2, verb contributions were minimal for the non-match pairs (PG image with SG caption), indicating that the model does not consider verbs as significant predictors. The FLAVA model incorrectly attributed high importance to SG verbs in PG image scenarios. Even large decoder VLMs like LLaVA-NeXT show significant shortcomings. The closest performance to human annotation is obtained by GPT-4o, however we can not study the results of this model in depth as the model's weights are not publicly available.

Overall, our findings highlight that current publicly available VL models, including those analyzed in this study, have significant room for improvement in integrating social and world knowledge information in grounding verb phrases that describe scenarios of different kinds.
These results underscore the need for advanced methodologies in training and evaluating VL models to enhance their ability to process and ground context accurately and integrate world knowledge, leading to more robust and reliable performance.






\section*{Acknowledgements}
This research was partially supported by \textit{DisAI - Improving scientific excellence and creativity in combating disinformation with artificial intelligence and language technologies}, a project funded by Horizon Europe under \href{https://doi.org/10.3030/101079164}{GA No. 101079164}; by the \textit{MIMEDIS}, a project funded by the Slovak Research and Development Agency under GA No. APVV-21-0114. The collaboration was facilitated by the Multi3Generation COST Action CA18231. The authors thank their anonymous reviewers for constructive comments on an earlier version of this paper.

\bibliographystyle{apacite}

\setlength{\bibleftmargin}{.125in}
\setlength{\bibindent}{-\bibleftmargin}

\bibliography{CogSci_Template}

\end{document}